\documentclass{article}
\usepackage{PRIMEarxiv}
\usepackage[numbers,sort&compress,sectionbib]{natbib}
\usepackage[utf8]{inputenc} 
\usepackage[T1]{fontenc}    
\usepackage{hyperref}       
\usepackage{url}            
\usepackage{booktabs}       
\usepackage{amsfonts}       
\usepackage{nicefrac}       
\usepackage{microtype}      
\usepackage{lipsum}
\usepackage{fancyhdr}       
\usepackage{graphicx}       
\graphicspath{{media/}}     
\usepackage[table]{xcolor}
\usepackage{colortbl} 
\usepackage{booktabs}
\usepackage{tabularx}
\usepackage{float}
\usepackage{amsmath,amssymb,amsfonts}
\usepackage{algorithmic}
\usepackage{graphicx}
\usepackage{textcomp}
\usepackage{pifont}
\usepackage{rotating}
\usepackage{xcolor}
\newcommand{\xmark}{\ding{55}}%
\title{A Hybrid Convolutional Neural Network with Meta Feature Learning for Abnormality Detection in Wireless Capsule Endoscopy Images}

\author{
	Samir Jain, Aparajita Ojha, Ayan Seal\\
	PDPM Indian Institute of Information Technology, Design and Manufacturing \\
	Jabalpur, India 482005\\
	\texttt{\{samirjain, aojha, ayan\}@iiitdmj.ac.in} \\
}

\begin{document}
\maketitle

\begin{abstract}
	Wireless Capsule Endoscopy is one of the most advanced non-invasive methods for the examination of gastrointestinal tracts. An intelligent computer aided diagnostic system for detecting gastrointestinal abnormalities like polyp, bleeding, inflammation, etc. is highly exigent in wireless capsule endoscopy image analysis. Abnormalities greatly differ in their shape, size, color, and texture, and some appear to be visually similar to normal regions. This poses a challenge in designing a binary classifier due to intra-class variations. In this study, a hybrid convolutional neural network is proposed for abnormality detection that extracts a rich pool of meaningful features from wireless capsule endoscopy images using a variety of convolution operations. It consists of three parallel convolutional neural networks, each with a distinctive feature learning capability. The first network utilizes depthwise separable convolution, while the second employs cosine normalized convolution operation. A novel meta-feature extraction mechanism is introduced in the third network, to extract patterns from the statistical information drawn over the features generated from the first and second networks and its own previous layer. The network trio effectively handles intra-class variance and efficiently detects gastrointestinal abnormalities. The proposed hybrid convolutional neural network model is trained and tested on two widely used publicly available datasets. The test results demonstrate that the proposed model outperforms six state-of-the-art methods with 97\% and 98\% classification accuracy on KID and Kvasir-Capsule datasets respectively. Cross dataset evaluation results also demonstrate the generalization performance of the proposed model.
\end{abstract}


\section{Introduction}
Wireless Capsule Endoscopy (WCE) is one of the most widely accepted technology for the examination of gastrointestinal (GI) tracts since its emergence in the year 2000 \cite{tacheci2008kapslova}. There are a variety of disorders in the GI tract like ulcers, bleeding, inflammatory, polyps, and tumors. To detect and diagnose these GI disorders, an endoscopy procedure captures a sequence of images of the GI region for the examination. However, the procedure brings discomfort to patients as a camera fitted endoscopy tube is inserted into the patient's body through mouth or rectum. WCE has gained popularity by examining deeper GI regions where conventional endoscopy tubes cannot reach. A WCE capsule keeps on moving in the GI tract and records its video that normally has more than $80,000$ frames \cite{koulaouzidis2015optimizing}. This poses a challenge in the manual inspection of the entire tract, as there may be very few abnormal frames as compared to normal frames. Further, the size of an abnormal region can be very small in a frame that may get missed. This requires a constant attention of human experts for long hours. Due to these challenges, Computer-Aided Diagnostic (CAD) tools have become a popular choice in the automatic detection of abnormal candidate frames for final examination by gastro-experts.

In the last decade, machine learning (ML) algorithms were prominent for the detection of GI abnormalities like bleeding \cite{yuan2015bleeding}, ulcer \cite{yu2012ulcer}, polyps \cite{yuan2015improved}, etc. Deep learning (DL) models employing convolutional neural networks (CNNs) also enticed the interest of researchers in the area of medical imaging due to their inherent capability of image feature extraction \cite{iakovidis2018detecting}. Although DL algorithms automatically extract patterns in the data, the main challenge is the scarcity of data. Small datasets with high class imbalance, large variations in size, shape, and texture of abnormalities in WCE images make it  difficult to build an efficient model \cite{xing2020zoom}. Also, if the GI anomalies are very small as compared to the frame size, their features may get dominated by the features of normal regions, that in turn affects the prediction.

In this study, a DL based classification model is proposed using a hybrid CNN architecture with three parallel CNNs. To deal with large variations in different types of anomalies, and to tackle the problem of visual similarity of certain anomalies with the normal regions in WCE images, two different convolution (Conv) operators are employed in each CNN. While the first CNN uses a cosine normalized Conv (CNC) in its Conv layers, \cite{luo2018cosine} the second CNN exploits a depth-wise separable Conv (DSC)  \cite{chollet2016xception}. The third CNN is a meta feature extraction network that works on the statistical patterns of the feature maps generated by Conv layers of the first and the second CNN and its own previous layer. The CNN-trio works independently in unison to gather distinguishing features of different types of abnormalities, which significantly improves the classification performance. The model also uses an attention layer \cite{wang2017residual} that tunes the salient regions in the input image before it is passed on to the CNNs and also accelerates the convergence. Highlights of the proposed DL model are summarized as follows.

    
    $\bullet$ A hybrid convolutional neural network model is introduced for abnormality detection in wireless capsule endoscopy images. The model consists of two feeder networks with distinct convolution operations and a novel meta-feature extraction network for improved classification performance. 
    
    $\bullet$ The first feeder network utilizes cosine normalized convolution to control the variance within a feature map that speeds-up the learning process, while the second network exploits depthwise separable convolution to learn richer representations with less number of parameters. 
    
    $\bullet$ The meta-feature extraction network derives a rich pool of meaningful features drawn over the statistical patterns of the feature maps generated by the feeder networks layer-by-layer.
	
	$\bullet$ An attention layer exploiting the residual attention mechanism is used as a pre-processing layer to generate saliency of abnormal regions in an image before it is passed on to the three networks.

	$\bullet$ The proposed model outperforms six state-of-the-art gastrointestinal abnormality detection methods on two publicly available datasets. Cross-dataset validation results also exhibit the proposed model's generalization capability.

The rest of the paper is organized as follows. Section 2 presents a brief overview of some of the related works. Section 3 presents the proposed hybrid CNN model for abnormality detection in WCE images. Datasets, experimental results, and analysis are detailed in Section 4, and Section 5 concludes the proposed study.

\section{Related Work}
Detection of pathology through vision-based CAD systems has been in place for quite some time. Early works on WCE video analysis were based on conventional ML methods that utilized color and texture based handcrafted features for the classification of WCE abnormalities. Local binary pattern (LBP), color histogram, \cite{iakovidis2014automatic}, statistical features like mean, standard deviation, skew, and energy \cite{sainju2013bleeding} were extensively used by researchers for WCE image classification. These methods were primarily designed for specific type of anomalies like bleeding, polyp, and ulcers.
\begin{figure}[!th]
	\includegraphics[scale=0.4]{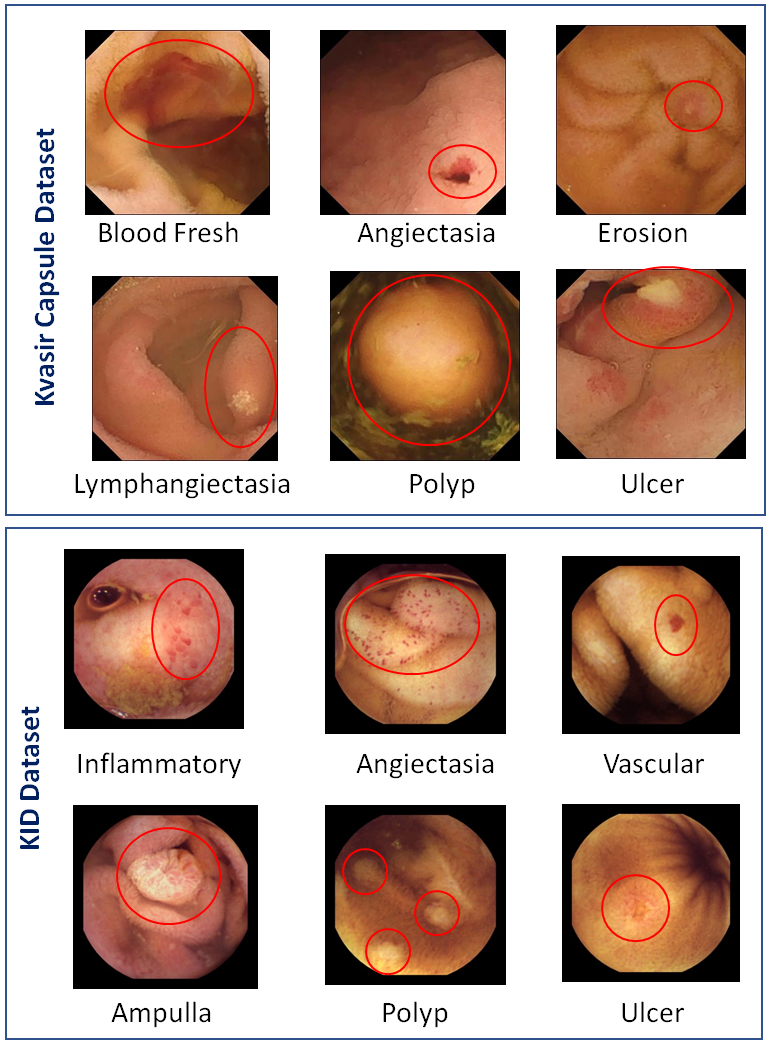}
	\centering
	\tiny
	\caption{WCE image samples with encircled abnormal regions.}
	\label{fanomaly}
\end{figure}

In recent years, CNNs have been quite effective in automatic extraction of generic features in WCE images. One of the first works on WCE image classification using a CNN can be found in Jia et al.\cite{jia2016deep}. They have used a 5-layer CNN for feature extraction and bleeding detection in WCE images. Since then, several CNN architectures have been proposed  for different types of abnormality detection \cite{georgakopoulos2016weakly}. Segui et al. \cite{segui2016generic} have used the Hessian and Laplacian of images along with the original RGB inputs to learn the features of WCE images using three parallel CNNs. Then the three CNN's features are fused and passed on to fully connected layers for image classification. Iakovidis et al. \cite{iakovidis2018detecting} have also employed a 5-layer CNN for anomaly detection and localization. The feature maps are utilized for the identification of salient points that may fall in the anomalous regions by using deep saliency detection and iterative cluster unification algorithms. Goel et al. \cite{goel2022dilated} have employed a two branch CNN, one as backbone network, and the other with resolution preserving dilated Conv operation. The backbone CNN extracts multiple features at different scales. The dilated Conv branch increases receptive field and helps in the extraction of dominant features. Finally, the features of backbone CNN and the dilated CNN are fused to fetch the global dominant features \cite{goel2022dilated}. 

Transfer learning approaches are commonly applied to train CNNs when the training data is limited. This situation is quite relevant to the existing WCE datasets and researchers have suggested the use of pretrained CNN architectures like AlexNet, GoogLeNet, ResNet, VGG16, etc. for the classification of WCE images \cite{ ayyaz2022hybrid}. Caroppo et al. \cite{caroppo2021deep} have contemplated the technique for feature extraction from intermediate layers of pretrained ResNet50, VGG16, and InceptionV3 models. The features were concatenated and optimized with the maximum relevance-minimum redundancy algorithm on which ML classifiers were trained to identify the bleeding images. Lan et al. \cite{lan2019deep} have also suggested the use of a pretrained VGG-16 network with region proposals to classify and localize abnormalities. Rustam et al. \cite{rustam2021wireless} have exploited MobileNet architecture in a custom-built CNN, for the classification of bleeding frames. These research articles signify the efficiency of deep transfer learning in the classification and localization tasks in WCE images. 

Size, shape, and texture of abnormal regions in WCE images vary in different types of abnormalities. In some images, an abnormal region could be very small in size. Further, the color and texture of an abnormality in some cases may be similar to that of a normal region. Figure \ref{fanomaly} shows these variations for different abnormalities like polyp, vascular, and inflammatory legions. In some cases, features of normal regions may dominate those of tiny abnormal regions. To deal with these challenges, attention-guided CNNs have been used by researchers. Xing et al.\cite{xing2020zoom} have suggested a novel two stage attention mechanism for zooming into abnormal regions that helps in detecting small abnormalities present in images. Jain et al. \cite{jain2021deep} have used squeeze-and-excitation blocks in a CNN for the identification of anomalies that attained higher accuracy.  
 
To deal with intra-class variations, Yuan et al. \cite{yuan2018riis} have introduced an image similarity constraint in the loss function to train a DenseNet for polyp identification in WCE images. A densely connected CNN with unbalanced discriminant loss and category sensitive loss was suggested by Yuan et al. \cite{yuan2019densely}, that helped in low intra-class variation and high inter-class variations. Guo et al. \cite{guo2019triple} have made a triple ANET architecture with angular contrastive loss with an attention mechanism to minimize intra-class variance for polyp detection. Methods available in the literature have mostly dealt with intra-class variation with a single type of abnormality. To deal with different types of anomalies while classifying them as abnormal, special mechanism needs to be designed that can capture the inherent characteristics of these abnormalities. Harnessing the power of different types of Conv operations, a parallel CNN architecture is proposed in the present paper that employs a novel meta-feature extraction technique to distinguish between a wide range of abnormalities using the statistical patterns hidden in the feature maps. These meta-features when clubbed with the conventional feature maps resulted in an improved performance of the classifier. In the next section, details of the proposed CNN model are presented. 
  \begin{figure*}[!th]
	\includegraphics[scale=0.35]{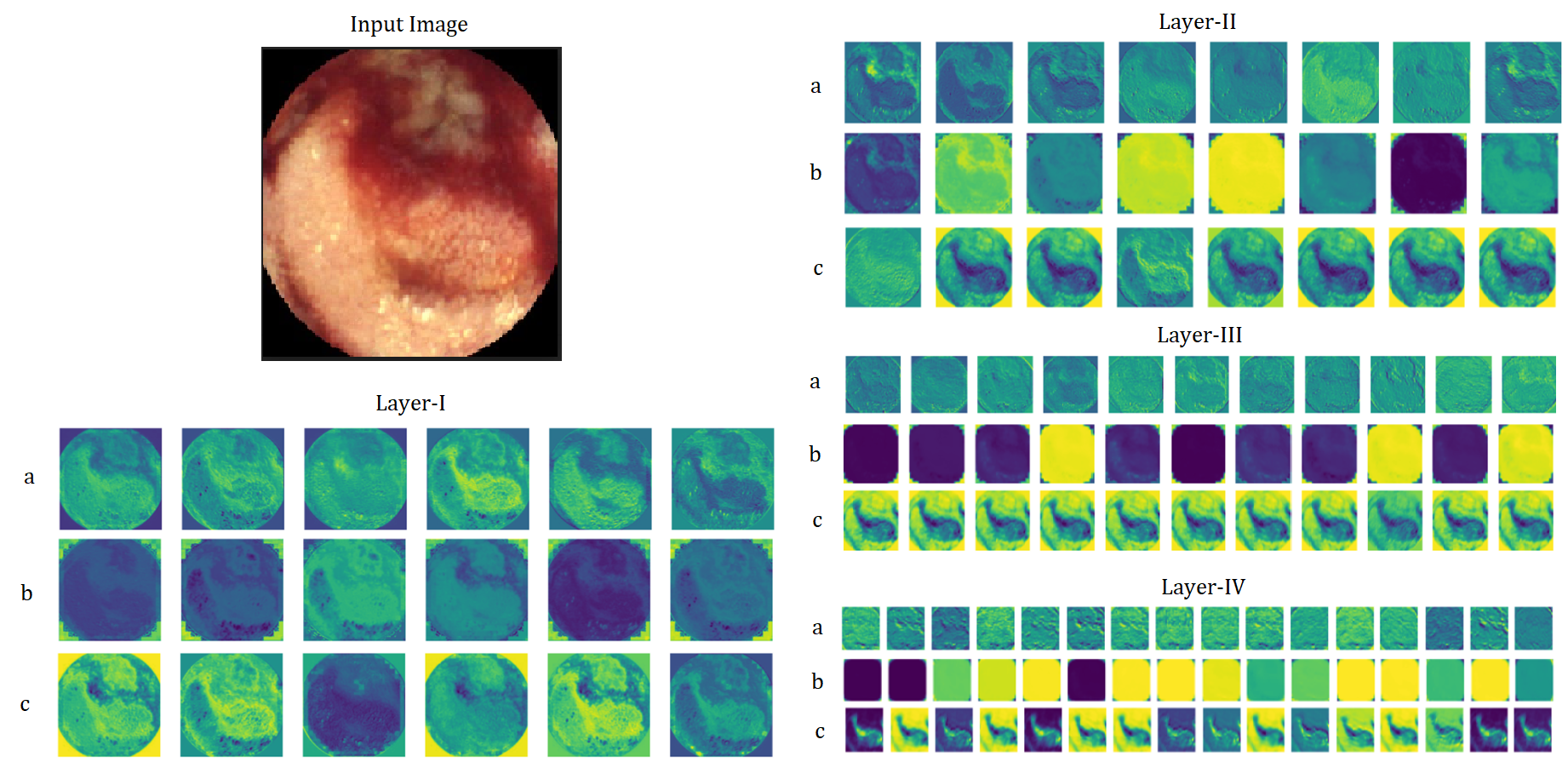}
	\centering
	\tiny
	\caption{Feature maps (FMs) extracted from the four layers (I, II, III, and IV) of the hybrid CNN network depicting different features learned by each CNN. (a) FMs from the DSC layer, (b) FMs from the CNC layer, and (c) FMs from the MFS layer.}
	\label{fgam}
\end{figure*}
\section{Methodology}

\begin{figure}[!th]
	\includegraphics[scale=0.60]{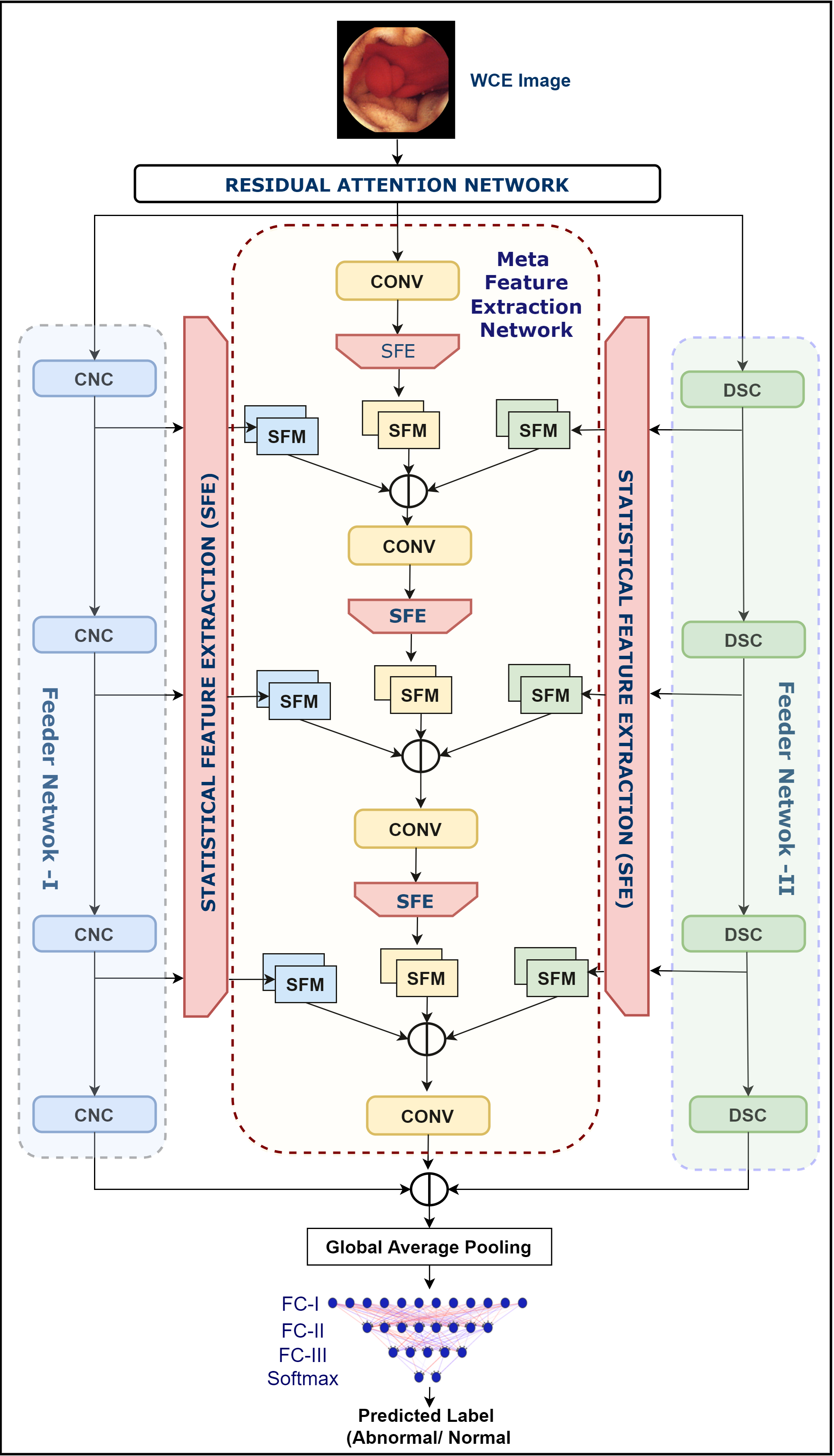}
	\centering
	\tiny
	\caption{Block diagram of hybrid CNN architecture adopted for the classification of WCE images.}
	\label{par-cnn}
\end{figure}

 To build a CNN model that can learn distinguishing features of WCE abnormalities, different types of Conv operations are utilized in the proposed DL model with three parallel CNNs. Before the input image is passed on to the CNN-trio, it is preprocessed through a residual attention mechanism \cite{wang2017residual} so that salient regions in images could be highlighted. The  first CNN uses CNC that works on cosine similarity in place of simple dot product. Using a CNC helps in controlling the variance within a feature map to clearly distinguish between the patterns arising from different feature maps of a Conv layer. This helps to speed-up the learning process. The second network exploits the DSC operation that is known to learn richer representations with a smaller number of parameters \cite{chollet2017xception}. Both these networks capture a diverse set of features, a sample of which can be seen in Fig. \ref{fgam}. With these two types of Conv operations, a rich pool of meaningful features can be extracted if their statistical patterns are analyzed as well. It may be noted that a large variance at a pixel position across all the feature maps may reveal saliency in pattern learning due to the diversity in the learned features. Further, pixel-wise maximum across all the feature maps is also an important statistic as it depicts the significance of the information at the particular location. In view of this, a novel meta-feature extraction (MFE) network is introduced that leverages the statistical information drawn over the features generated by Conv layers of the two previous CNNs that are feeder networks to the MFE network termed as FN-1 and FN-2. Pixelwise variance and maximum of the feature maps are computed at the output of each Conv block of the feeder networks, and are supplied to MFE network that makes use of a conventional Conv operation. Input to each Conv block of the MFE network is a hybrid set of the statistical features of its previous Conv block and the corresponding Conv blocks of the feeder networks. It then generates meta features from these statistical patterns of the feature maps. The architecture of proposed hybrid CNN model for WCE abnormality detection is depicted in Fig. \ref{par-cnn}. 

\begin{figure}[!h]
	\includegraphics [scale=0.5]{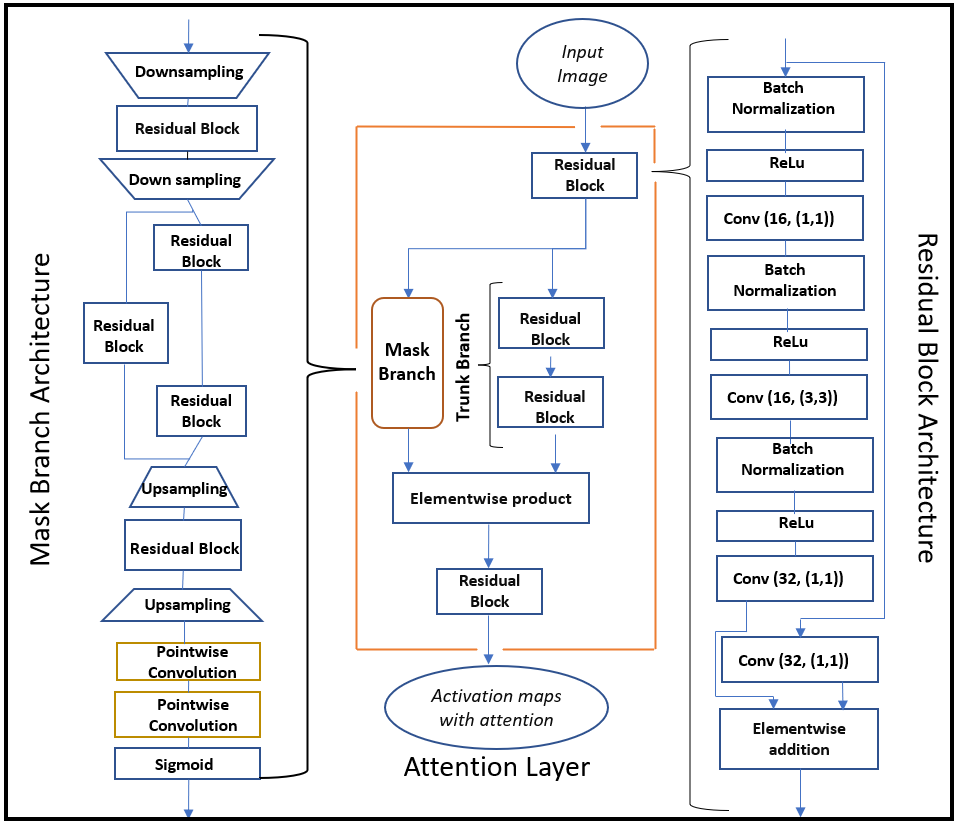}
	\centering
	\tiny
	\caption{Block diagram of the residual attention mechanism adopted in the proposed CNN.}
	\label{ran-fig}
\end{figure}

The first layer in the model is an attention layer that uses residual attention mechanism as given in Fig. \ref{ran-fig}. This layer works as a preprocessing layer for learning salient regions of possible abnormalities in the input images. The output of the attention layer is forwarded to FN-1, FN-2 and the MFE network. Each of these CNNs has $4$ Conv blocks. Each block contains a Conv layer with $3\times3$ filters followed by a batch normalization layer, a ReLU activation layer and a $2\times2$ max-pooling layer. In WCE images, anomalies can be minute in size and small filter size of $3 \times 3$ for Conv operation fits best to extract the relevant features at local level. The outputs of the fourth Conv blocks of all the three CNNs are  concatenated and are passed through a global average pooling layer that helps in preventing the model from overfitting \cite{lin2013network}.  Four fully connected (FC) layers are appended to the network after the GAP layer containing $1024$, $512$, $128$, and $2$ units subsequently. The first three layers use ReLU activation whereas the last one uses the softmax activation function. The network is trained to predict the label of the input image as normal or abnormal. 

The attention mechanism helps in the extraction of features related to small anomalies as they otherwise get dominated by the widely spread normal regions in images \cite{xing2020zoom}. The proposed CNN model utilizes a residual attention mechanism introduced by Wang et al. \cite{wang2017residual} that extracts attention-aware features as displayed in Fig. \ref{ran-fig}. It captures potential features from the input image to generate mask that suppresses the background responses. The attention layer is designed with the help of two branches, trunk branch and mask branch. The trunk branch makes use of multiple residual blocks for feature extraction. The mask branch generates a mask that suppresses the background regions.

\begin{figure}[!h]
	\includegraphics[scale=0.4]{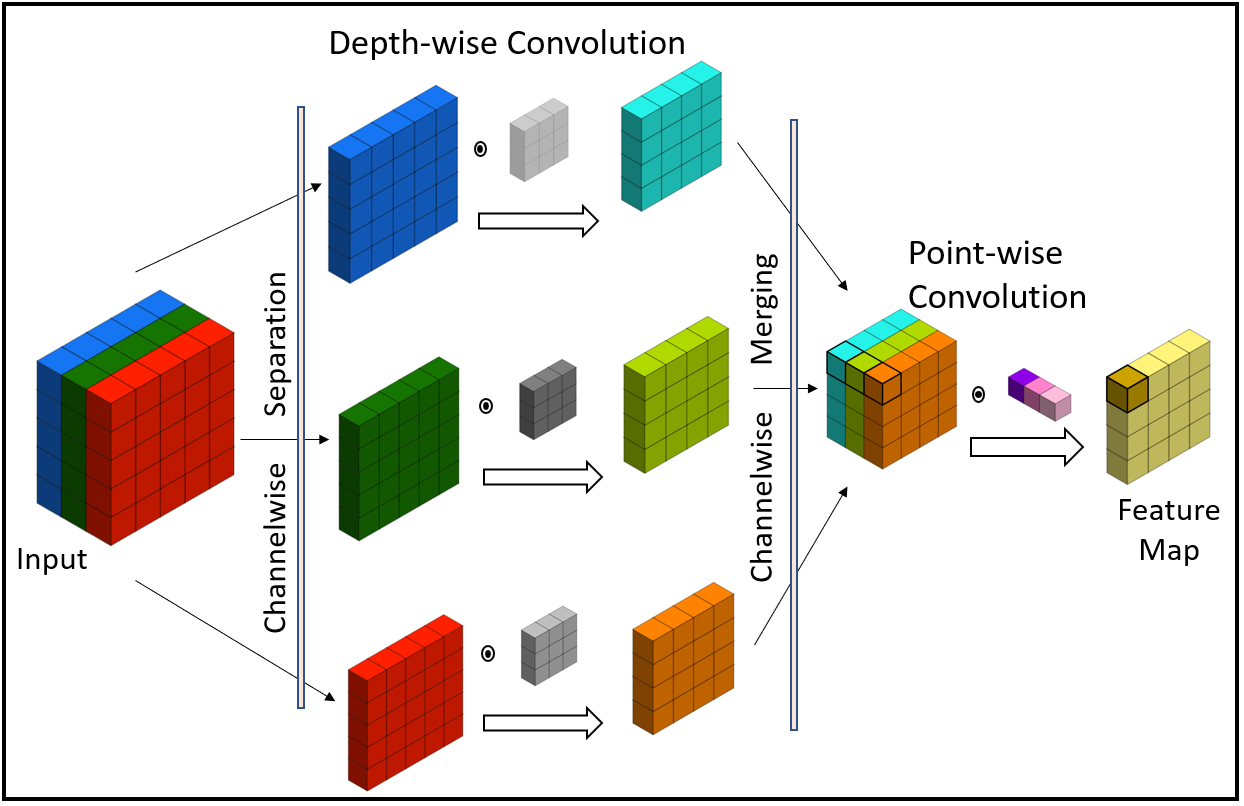}
	\centering
	\tiny
	\caption{Depthwise separable convolution mechanism.}
	\label{fg2}
\end{figure}

As discussed earlier, the proposed CNN model utilizes three different types of convolutions. Conventionally, a Conv operation performs the dot product between an input and its filter (weight) vector. The activation on the result of the dot product may lead to large variations, making the network sensitive to small changes in the input distribution. In a CNC \cite{luo2018cosine}, the dot product is replaced by cosine similarity that bounds the output in the range $[-1,1]$ as given in Eq. \ref{cnc_2}.

\begin{equation}\label{cnc_2}
	o=f(\frac{w \cdot x}{\vert w \vert \cdot \vert x \vert}),  
\end{equation}
where $w$ is the weight vector and $x$ is the input vector. In the hybrid CNN the FN-1 exploits CNC. Further, DSC is used by FN-2 (cf. Fig. \ref{par-cnn}). The DSC introduced by Chollet et al. \cite{chollet2016xception} involve two consecutive Conv operations, the depthwise Conv and the pointwise Conv. In the depthwise Conv, spatial Conv is performed independently on each channel of the input. Pointwise Conv is a $1\times 1$ Conv operation that projects the channel output from depthwise Conv into a new channel space. A pictorial view of the DSC is presented in Fig. \ref{fg2}. The DSC helps the network train faster by reducing the number of multiplications as compared to a conventional Conv operation.

\begin{figure}[!h]
	\includegraphics[width=\linewidth]{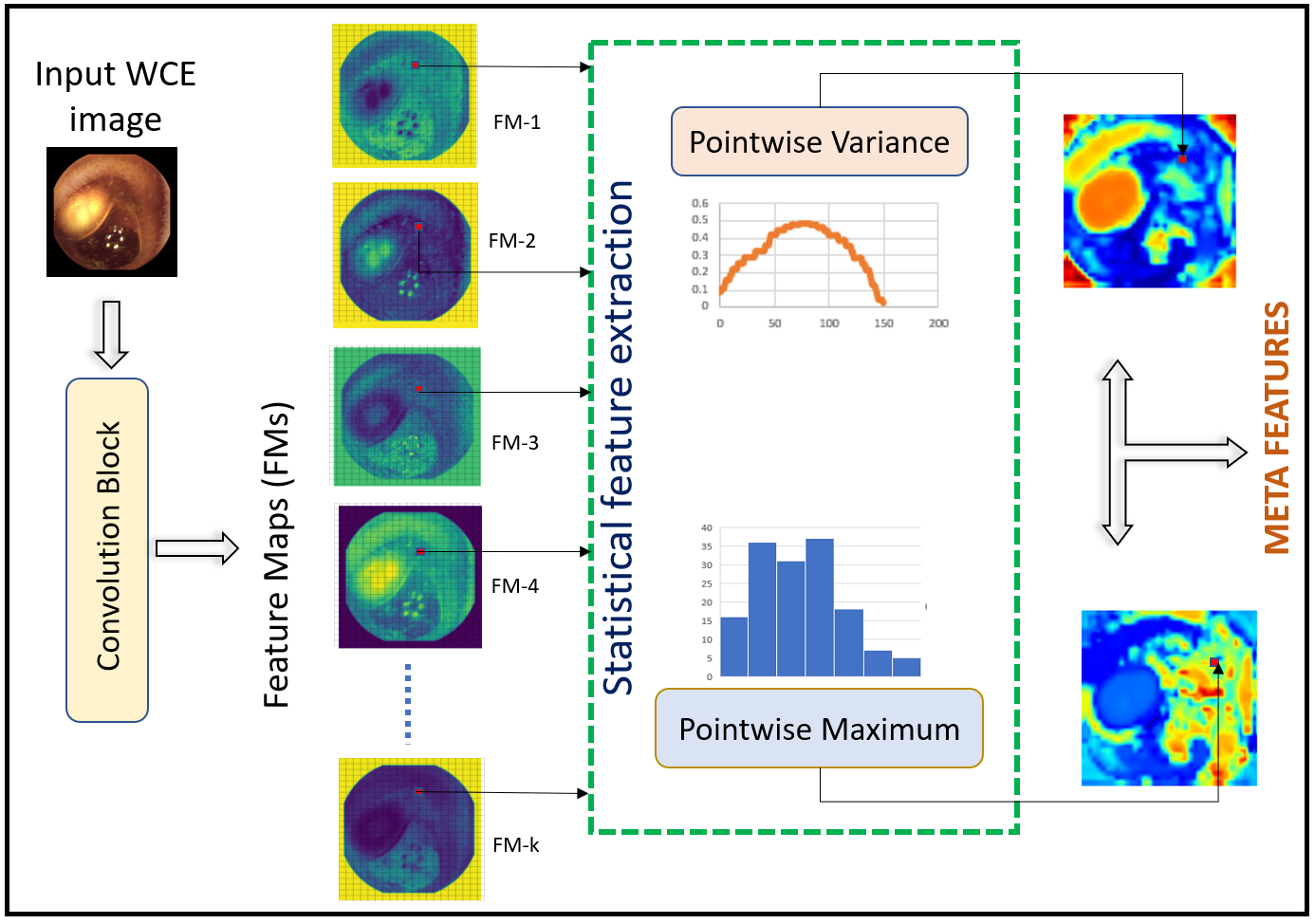}
	\centering
	\tiny
	\caption{Process of the proposed meta-feature extraction layer.}
	\label{mfs-layer}
\end{figure}

As mentioned earlier, statistical feature maps (SFMs) are generated by taking the pixelwise variance and maximum of all the feature maps generated through each Conv block of FN-1, FN-2, and MFEN. Let there be $k$ feature maps at the output of the $i-th$ Conv block given by $F_i = \{ F_i^j \}_{j=1}^k$, each of the size $n \times m$. Then two SFMs are computed using the variance and maximum at each pixel location $(p,q)$ across all the feature maps, as given in Eqs. \ref{mfs1} and \ref{mfs2}. 
\begin{equation}\label{mfs1}
	V_{F_{i}}(p,q) =maximum\{F_{i}^{j}(p,q)|j=1,2,3,.....k\}           
\end{equation}
\begin{equation}\label{mfs2}
	M_{F_{i}}(p,q) = variance\{F_{i}^{j}(p,q)|j=1,2,3,.....k\}         
\end{equation}   

Thus, each output of FN-1 and FN-2 produces two feature maps that are fed to the MFE network. The initial Conv block of the MFE network takes the same input as the other two networks which is the output of the attention layer. Subsequent Conv blocks of the MFE network work on a hybrid of the SFMs of the feeder networks and SFMs of its own feature maps generated by the previous Conv block. The main idea behind using the variance and maximum is that the textural and color components of the input may express the image characteristics in a more efficient way as shown in Fig. \ref{mfs-layer}. 

\section{Experimental results and discussion}
In this section, the training details and the performance of the proposed hybrid CNN model are discussed. Starting with a description of the dataset, results of the performance of the proposed hybrid CNN and six other state-of-the-art methods on the datasets are presented. Results are analysed in detail and the efficacy of the proposed method is established. 

\subsection{Datasets}
Two popular and publicly available WCE image datasets are used in the proposed work. The datasets use different WCE apparatuses to capture images of GI regions. The first dataset is KID \cite{pmid28580415} which is termed as Dataset-I and the second dataset Kvasir-Capsule \cite{smedsrud2020} is referred as Dataset-II.
\subsubsection{Dataset -I (KID)}
This is a publicly available dataset containing color images and videos. For the present experiment, only images are considered. The dataset contains two subsets of endoscopic images. KID Dataset-1 \cite{iakovidis2014automatic} and KID Dataset-2 \cite{iakovidis2018detecting}. KID Dataset-1 contains $77$ abnormal samples belonging to angioectasias (27), aphthae (5), chylous cysts (8), polypoid lesions (6), villous oedema (2), bleeding (5), lymphangiectasias (9), ulcers (9), and stenoses (6). KID Dataset 2\cite{iakovidis2018detecting}, contains $2371$ images of variegated findings of polypoid (44), vascular (303), inflammatory lesions (227), and ampulla-of-vater (19). It also contains assorted images belonging to normal regions of the esophagus (282), stomach (599), small bowel (728), and colon (169). Both the datasets are merged in the present work to have a total of $670$ abnormal samples and $1778$ normal samples. Each image has dimensions $360 \times 360$ with 3 channels.    	      
\subsubsection{Dataset-II (Kvasir-Capsule)}     
This dataset comprises $47,238$ labeled images \cite{smedsrud2020} with $14$ labels, $9$ labels of abnormal category and $4$ labels of normal category. There is another category called 'foreign-body' that is not considered in the study. The abnormal samples contain images of angiectasia (866), blood fresh (44), blood hematin (12), erosion (506), erythema (159), lymphangiectasia (92), polyp (55), ulcer (854). In the normal category, the images belong to the ileocecal valve (4,189), normal clean mucosa (34,338), pylorus (1,529), reduced mucosal view (2906). Images of all normal categories are combined into one label that is normal and all abnormal categories into another label termed as abnormal. In this way, there are $42,962$ normal images and $3,500$ abnormal images. All the samples are 3-channel RGB images with a size of $336 \times 336$. 

To train the hybrid CNN and other state-of-the-art models for comparison, both the datasets are divided into training and testing sets in the ratio of $80:20$. Due to relatively lower number of samples under the abnormal category a class imbalance problem exists in both the datasets, that affects the classification accuracy of the model. In Dataset-I, data augmentation is applied using image flipping, rotation, and the addition of Gaussian noise to balance the minority class. Since this set contains a smaller number of images, augmentation on both the categories is performed to produce a total of $3,000$ samples in each class. In the case of Dataset-II, sub-sampling is applied to the normal class, as there are a large of images in this class. From the normal class, $3,500$ number of images are sampled randomly for the present work.   
\subsection{Results and Comparison}
The proposed model is implemented and tested on  Tensorflow-GPU 1.15 platform with Keras 2.1.4 installed on a Windows 10 workstation with $32$ GB RAM that has Intel Xeon CPU and Nvidia Quadro P4000 8GB GPU. The proposed CNN is trained for $100$ epochs using the stochastic gradient descent optimizer with a learning rate of $0.001$. The categorical cross-entropy loss is used to train the network. The model is trained and tested on both Dataset-I and Dataset-II. An ablation study is also performed by removing and adding the attention layer, DSC, MFS, and  CNC layers. The effect of each module is measured to find out the best configuration for the proposed Hybrid CNN model. Results of the ablation study are recorded in Table \ref{ablation study}. Number of trainable parameters and FLOPS count of each configuration in the ablation study are also compared in Table \ref{flops}.

\begin{figure}[!th]
	\includegraphics[scale=0.40]{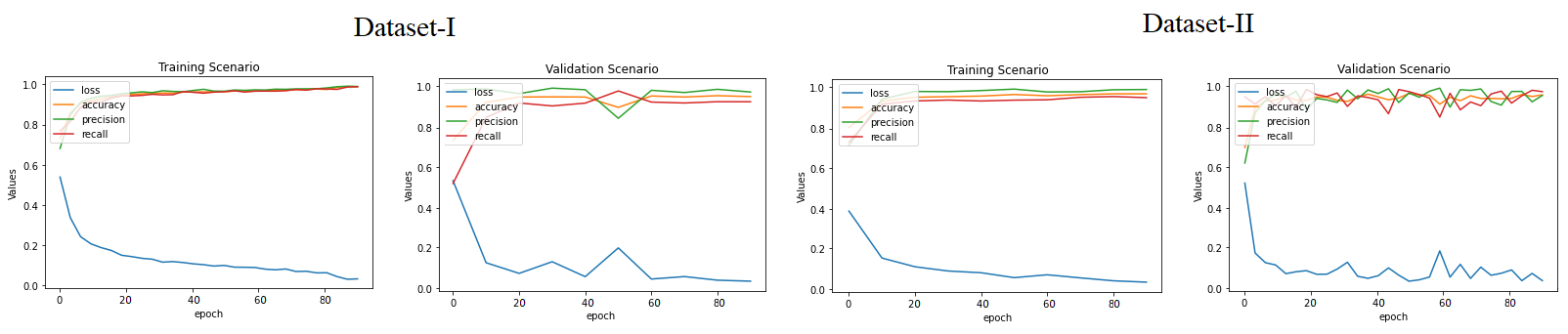}
	\centering
	\tiny
	\caption{Plot of training and validation accuracy, precision, recall, and loss curves of the hybrid CNN model on Dataset-I and Dataset-II.}
	\label{roc}
\end{figure}  
\begin{table}[!h]
	\centering
	\caption{Ablation studies performed on the hybrid CNN by adding and removing specific layers and blocks.}
	\label{ablation study}
	\scalebox{0.8}{
		\begin{tabular}{c c c c c c c c c c c} \hline
			\textbf{Dataset} & \textbf{Attention} & \textbf{DSC} & \textbf{MFE} & \textbf{CNC} &  \textbf{AC} & \textbf{PR} & \textbf{RE}  & \textbf{F1} \\  \hline
			Dataset-I & \checkmark & \checkmark & \checkmark & \checkmark &\textbf{95.45} & \textbf{95.00} & 95.86 & \textbf{95.42}\\
			&\xmark & \checkmark&\checkmark&\checkmark&94.90&93.57&\textbf{96.24}&94.89\\
			&\checkmark & \checkmark&\xmark&\checkmark&94.57&94.04&96.00&95.10\\
			&\checkmark & \xmark&\xmark&\checkmark&92.65&92.57&94.01&93.28\\
			&\checkmark & \checkmark &\xmark&\xmark&92.38&95.11&90.07&92.52\\
			&\xmark & \checkmark&\xmark&\checkmark&93.28&90.84&93.39&92.10\\
			&\xmark & \xmark&\xmark&\checkmark&91.76&95.04&92.28&93.64\\
			&\xmark & \checkmark&\xmark&\xmark&88.53&86.33&91.40&88.79\\\hline
			Dataset-II & \checkmark & \checkmark & \checkmark & \checkmark &\textbf{97.38} & \textbf{97.38} & \textbf{97.38} & \textbf{97.38}\\
			&\xmark & \checkmark&\checkmark&\checkmark&96.14&96.60&96.27&96.43\\
			&\checkmark & \checkmark&\xmark&\checkmark&95.42&94.85&95.95&95.40\\
			&\checkmark & \xmark&\xmark&\checkmark&92.17&95.04&92.22&93.61\\
			&\checkmark & \checkmark &\xmark&\xmark&92.42&88.85&95.69&92.14\\
			&\xmark & \checkmark&\xmark&\checkmark&94.14&95.71&96.43&96.07\\
			&\xmark & \xmark&\xmark&\checkmark&90.11&91.54&90.07&90.80\\
			&\xmark & \checkmark&\xmark&\xmark&91.46&92.00&92.28&92.14\\ \hline
		\end{tabular}}
\end{table}
\begin{table}[!h]
	\centering
	\caption{Comparison of trainable parameters and flops count on the proposed CNN by adding and removing specific layers and blocks.}
	\label{flops}
	\scalebox{0.8}{\begin{tabular}{c c c c c c c c c } \hline
			\textbf{Attention} & \textbf{DSC} & \textbf{MFS} & \textbf{CNC} &  \textbf{\# Parameters} & \textbf{\# Flops} \\
			&  &  &  &  \textbf{(million)} & \textbf{(giga)}\\
			\hline
			\checkmark & \checkmark & \checkmark & \checkmark &1.94&14.0\\
			\xmark & \checkmark&\checkmark&\checkmark&1.58&11.0\\
			\checkmark & \checkmark&\xmark&\checkmark&1.32&5.37\\
			\checkmark & \xmark&\xmark&\checkmark&0.99&4.9\\
			\checkmark & \checkmark &\xmark&\xmark&0.69&1.78\\
			\xmark & \checkmark&\xmark&\checkmark&1.21&3.16\\
			\xmark & \xmark&\xmark&\checkmark&0.96&2.81\\
			\xmark & \checkmark&\xmark&\xmark&0.64&0.36\\ \hline
	\end{tabular}}
\end{table}
\begin{figure}[!h]
	\includegraphics[scale=0.35]{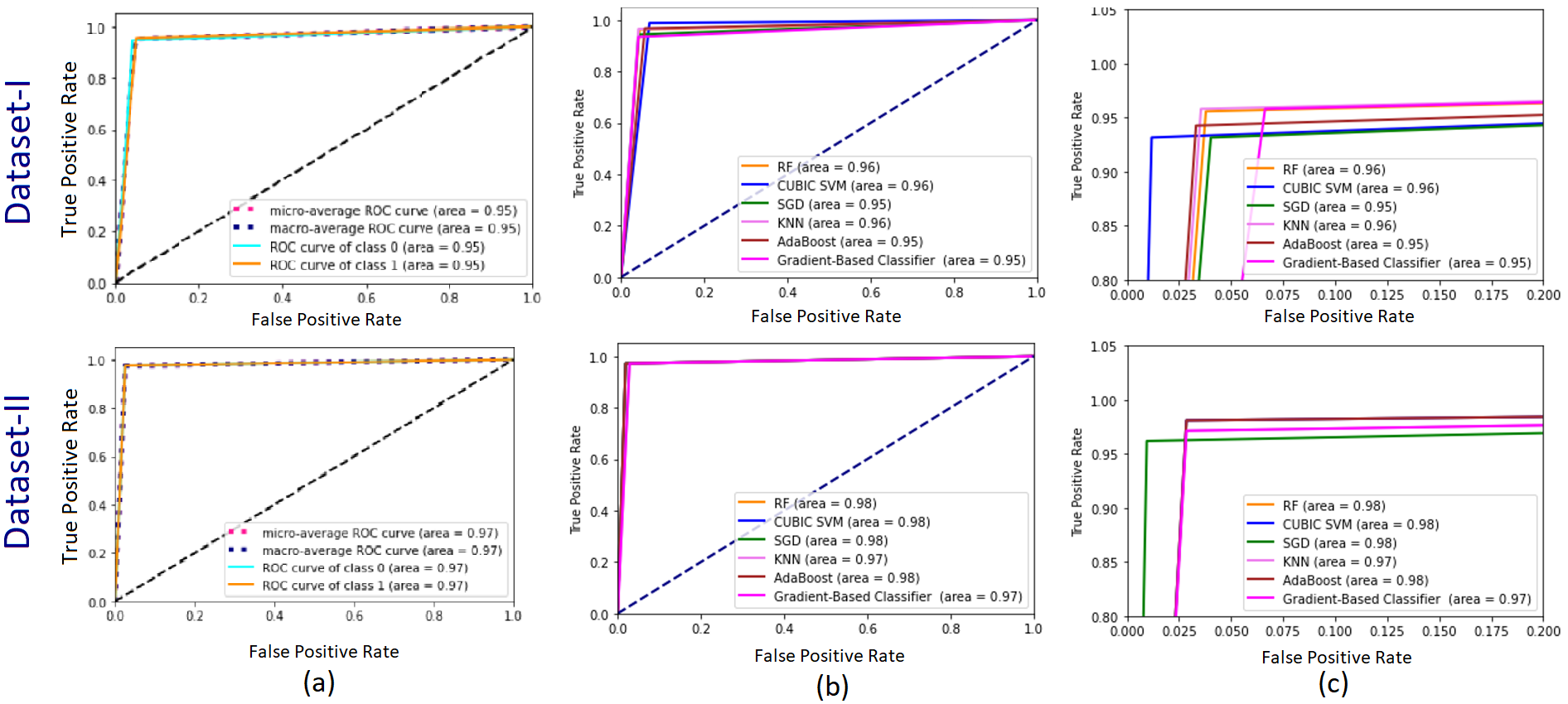}
	\centering
	\tiny
	\caption{Plot of ROC curves on classification of test samples from Dataset-I and Dataset-II. (a) From hybrid CNN, (b) Using ML methods trained on features extracted from hybrid CNN, (c) The magnified version of b.}
	\label{roc}
\end{figure}
The outcome of ablation study shows that the best results are achieved with the utilization of attention layer, DSC, CNC and MFS layers on both the datasets. The highest accuracy (AC) and F1-score (F1) are  $95.45\%$ and $95.42\%$ respectively for Dataset-I. For Dataset-II  both the scores are $97.38\%$. The precision (PR) and recall (RE) are found to be $95.00\%$ and $95.86\%$ for Dataset-I whereas $97.38\%$ PR and RE, are recorded for Dataset-II. The impact of MFS layer on the performance of the proposed hybrid CNN can be seen as the accuracy falls when the MFS is not utilized. When the network is built with the DSC alone it gives the lowest AC of $88.53\%$ on Dataset-I whereas the proposed network performs weakest on Dataset-II with $90.11\%$ AC when configured with the CNC alone. This suggests that the CNC and DSC complement each other in recognizing the image features. Taking the best configuration of the proposed CNN, ROC curves are also plotted that record an AUC of $0.95$ on Dataset-I and $0.97$ for Dataset-II (Fig. \ref{roc}).

In this study, the feature extraction capability of the proposed CNN is also assessed. This experiment uses the penultimate layer of the trained CNN model to analyze its feature extraction capability. Features of length $128$ are extracted from this layer and are fed to some of the frequently used ML algorithms for binary classification task. These algorithms include Support Vector Machines (SVM) \cite{jain2020detection}, Random Forest (RF) \cite{jain2020detection}, AdaBoost (AB) \cite{schapire2013explaining}, Stochastic Gradient Descent (SGD) \cite{camp2022supervised}, and Gradient Boosting Classifier (GBC) \cite{dimililer2022dct}. The results of the classifiers trained using the CNN features are listed in Table \ref{ML_classifiers} for both the datasets. The RF classifier is configured with 100 estimators where Gini index is utilized to assess the quality of split. SVM with a cubic kernel is exploited wherein regularization parameter is set to '1' and 'L2' penalty is applied. SGD is configured with a hinge loss and L2 regularizer whereas in GBC, logistic regression loss is used with 100 decision tree (DT) estimators. KNN is configured with 5 nearest neighbours and in AB 100 DT estimators are used. Train and test sets used in the experiment are the same as those for training the hybrid CNN. It can be inferred from the results that the RF classifier is performing consistently on both the datasets with the highest AC of $96.60\%$ on Dataset-I and $98.12\%$ on Dataset-II. RF is an ensemble classifier that exploits more than one decision tree (estimators) for final prediction. This can be the probable reason for the better performance of hybrid CNN as the feature extractor with RF. SVM performs very close to RF on both the datasets with an AC of $96.57\%$ on Dataset-I and $97.88\% $ on Dataset-II. The ROC curves of all the classifiers are plotted in Fig. \ref{roc}. The highest AUC of $96\%$ is attained on Dataset-I with RF, Cubic SVM and KNN classifiers while on Dataset-II it is recorded to be $98\%$ with RF, Cubic SVM, SGD, and AB. For the Dataset-I, SGD, AB, and GBC are performing marginally lower by 1\% and on Dataset-II, KNN and GBC have shown 1\% drop in the performance under AUC metric as compared to the other classifiers. The performance analysis of these ML classifiers suggests that all of them are performing considerably well. 
\begin{table}[!h]
	\centering
	\caption{Classification results of ML algorithms trained on the features extracted using the hybrid CNN with 5-fold cross-validation.}
	\label{ML_classifiers}
	\scalebox{0.8}{\begin{tabular}{c c c c c c} \hline
			\textbf{Dataset} & \textbf{Classifier} &\textbf{AC} & \textbf{PR} & \textbf{RE}  & \textbf{F1}\\ \hline
			Dataset-I &RF&\textbf{96.60}&\textbf{97.89}&97.78&\textbf{97.83}\\
			&SVM&96.57&97.79&97.57&97.68\\
			&SGD &94.78&92.03&\textbf{97.89}&94.87\\ 
			&KNN&95.05&95.28&94.60&94.94\\
			&AB&95.29&94.01&96.44&95.21\\
			&GBC&94.84&94.19&94.88&94.53\\
			\hline
			Dataset-II &RF&\textbf{98.12}&\textbf{97.91}&\textbf{98.93}&\textbf{98.34}\\
			&SVM&97.88&97.89&96.48&97.19\\
			&SGD &96.42&96.51&96.33&96.42\\ 
			&KNN&97.00&96.79&96.77&96.78\\
			&AB&96.71&96.29&96.92&96.60\\
			&GBC&95.28&95.26&94.74&95.01\\
			\hline
	\end{tabular}}
\end{table}

The performance of the hybrid CNN is also compared with six state-of-the-art methods relevant to this study. These methods include the works by Goel et al. \cite{goel2022dilated}, Caroppo et al.\cite{caroppo2021deep}, Patel et al. \cite{patel2020automated}, Jain et al. \cite{jain2020detection}, Iakovidis et al. \cite{iakovidis2018detecting}, and Jia et al. \cite{jia2016deep}. Goel et al. \cite{goel2022dilated} have devised a resolution preserving multi-scale CNN based on dilated Conv operations. The CNN comprises  of 2 conventional Conv layers and 3 dilated Conv layers in parallel. Caroppo et al. \cite{caroppo2021deep} have employed 3 pretrained parallel CNNs, InceptionV3, ResNet50, and VGG16 for feature extraction. The extracted features of the three CNNs are  combined and optimized using the maximum relevance minimum redundancy (MRMR) algorithm for classification of WCE frames. Patel et al. have suggested an ML-based model employing dictionary learning with sparse encoding \cite{patel2020automated} using Local Binary Pattern (LBP) features around SIFT key points in WCE images. Their model uses the SVM to classify WCE bleeding images. Jain et al. \cite{jain2020detection} have suggested bringing out fractal features using the differential box-counting method to train the RF classifier. Iakovidis et al. \cite{iakovidis2018detecting} have devised an 8-layer CNN for classifying WCE images into the normal and abnormal categories. Jia et al. \cite{jia2016deep} have employed a 5-layer CNN for the identification of bleeding images. The best performing parametric configuration of each method as mentioned in the original research articles,  is used in the experiment. These parameters are listed in Tables \ref{SOTA_dl_pc} and \ref{SOTA_ml_pc} separately based on the type of the method, whether DL based or conventional ML method. In Table. \ref{SOTA_dl_pc} CL denotes Conv layers, PL is pooling layer, FCL refers to fully connected layer. Under the hyperparameters column, opt denotes the optimizer, lr is the learning rate, bs is the batch size and ep refer to number of epochs. TRP in the last column denotes a total count of trainable parameters.  
\begin{table}[!h]
    	\centering
    	\caption{\textup{Parametric configuration of three state-of-the-art CNN classification models and the proposed Hybrid CNN classifier.}}
    	\label{SOTA_dl_pc}
    	\scalebox{0.8}{\begin{tabular}{cccccc} \hline
    			\textbf{Author} & \textbf{\# CL} &\textbf{\# PL}  & \textbf{\# FCL} & \textbf{Hyperparameters} & \textbf{\# TRP}\\ 
    			& & & &\textbf{(opt, lr, bs, ep)} & \\
    			\hline
    			Jia et al. \cite{jia2016deep} & $3$ & $3$&$2$&SGD, 0.001, 16, 100 & 10,697,060\\
    			Iakovidis et al.\cite{iakovidis2018detecting} & $5$ & $4$ & $3$&SGD, 0.001, 16, 100 & 286,344 \\
				Caroppo et al.\cite{caroppo2021deep} & $152$ & $22$ & $5$&Adam, 0.001, 16, 100 & 60,752,386 \\
				Goel et al.\cite{goel2022dilated} & $7$ & $5$ & $3$&Adam, 0.001, 16, 100 & 278,722 \\
    			Hybrid CNN & $50$ & $12$ & $4$& SGD, 0.001, 16, 100 &1,943,096\\
    			\hline
    	\end{tabular}}
    	
    \end{table}
 \begin{table}[!h]
	\centering
	\caption{\textup{Parametric configuration of the ML classification methods compared with Hybrid CNN classifier.}}
	\label{SOTA_ml_pc}
	\scalebox{0.8}{\begin{tabular}{ccccc}\hline
			\textbf{Author}& \textbf{Method} & \textbf{Feature} &\textbf{ML}  & \textbf{Hyperparameters}\\ 
			& & \textbf{Length} &\textbf{Method}  & \\ 
			\hline
			Patel et al. \cite{patel2020automated} & SIFT+LBP+& 300 & SVM & cubic kernel (SVM),  \\
			
			& Sparse Coding &&& sparsity = 5,\\
			&  &&& dict size = 300  \\
			
			Caroppo et al. \cite{caroppo2021deep} & CNN+MRMR & 384 & SVM & cubic kernel \\
			Jain et al. \cite{jain2020detection}& DBC (FD) & 2025 & RF & $\#$ estimators = 500 \\
			Hybrid CNN & CNN & 128 & RF & $\#$ estimators = 100 \\
						\hline
	\end{tabular}}
\end{table} 
The comparative results of these methods and the hybrid CNN as listed in Table \ref{SOTA} demonstrate that the proposed CNN model outperforms other state-of-the-art methods on both the datasets. The second-best performance is exhibited by the model given by Caroppo et al. \cite{caroppo2021deep} and Goel et al. \cite{goel2022dilated}. When the proposed CNN is trained with a softmax layer and compared with \cite{goel2022dilated} and \cite{caroppo2021deep}, it shows better performance with an enhancement of $1\%$ F1 score on Dataset-I. It levels up F1 by 2\% on Dataset-II in comparison with \cite{caroppo2021deep} and 4\% with \cite{goel2022dilated}. Although the AUC of both \cite{caroppo2021deep} and the hybrid CNN are the same on Dataset-I, there is an increase of 1\% on Dataset-II. When the hybrid CNN is used as the feature extractor with the RF classifier, this configuration shows an improvement of $1\%$ in the AUC and 4\% in the F1 score for Dataset-I, and the performance steps up by $2\%$ on the AUC and $3\%$ in the F1 score on Dataset-II, as compared to the other two methods \cite{caroppo2021deep} and \cite{goel2022dilated}. Reliability of the proposed method is also validated using Cohen's Kappa score (CKS) where the proposed CNN scores better on both the datasets. It may also be noted that the statistical features from the feature maps generated by the feeder networks are shared with the MFE network that help in better learning of the features by the hybrid CNN model as compared to the model in \cite{caroppo2021deep} where three deep CNN architectures are trained independently for feature extraction with no feature sharing. Attention mechanism used as the preprocessing layer in the proposed model gives the added advantage in better extraction of features as  compared to \cite{caroppo2021deep} and \cite{goel2022dilated} that do not use any attention mechanism. This is also validated by the ablation study of the proposed model as shown in Table. \ref{ablation study}. In the case of dilated CNN proposed by Goel et al. \cite{goel2022dilated}, the CNN extracts global dominant features, which may misinterpret the class of abnormal frame having small sized anomaly or anomalies. The CNN models introduced by Iakovidis et al. \cite{iakovidis2018detecting} and Jia et al. \cite{jia2016deep} are shallow models (Table \ref{SOTA_dl_pc}) and that could be the reason for their relatively low performance. The handcrafted feature descriptors suggested by Jain et al. \cite{jain2020detection} and Patel et al. \cite{patel2020automated} might have limitations in capturing the diversity of features of different types of abnormalities in the datasets.

\begin{figure}[!h]
	\includegraphics[scale=0.50]{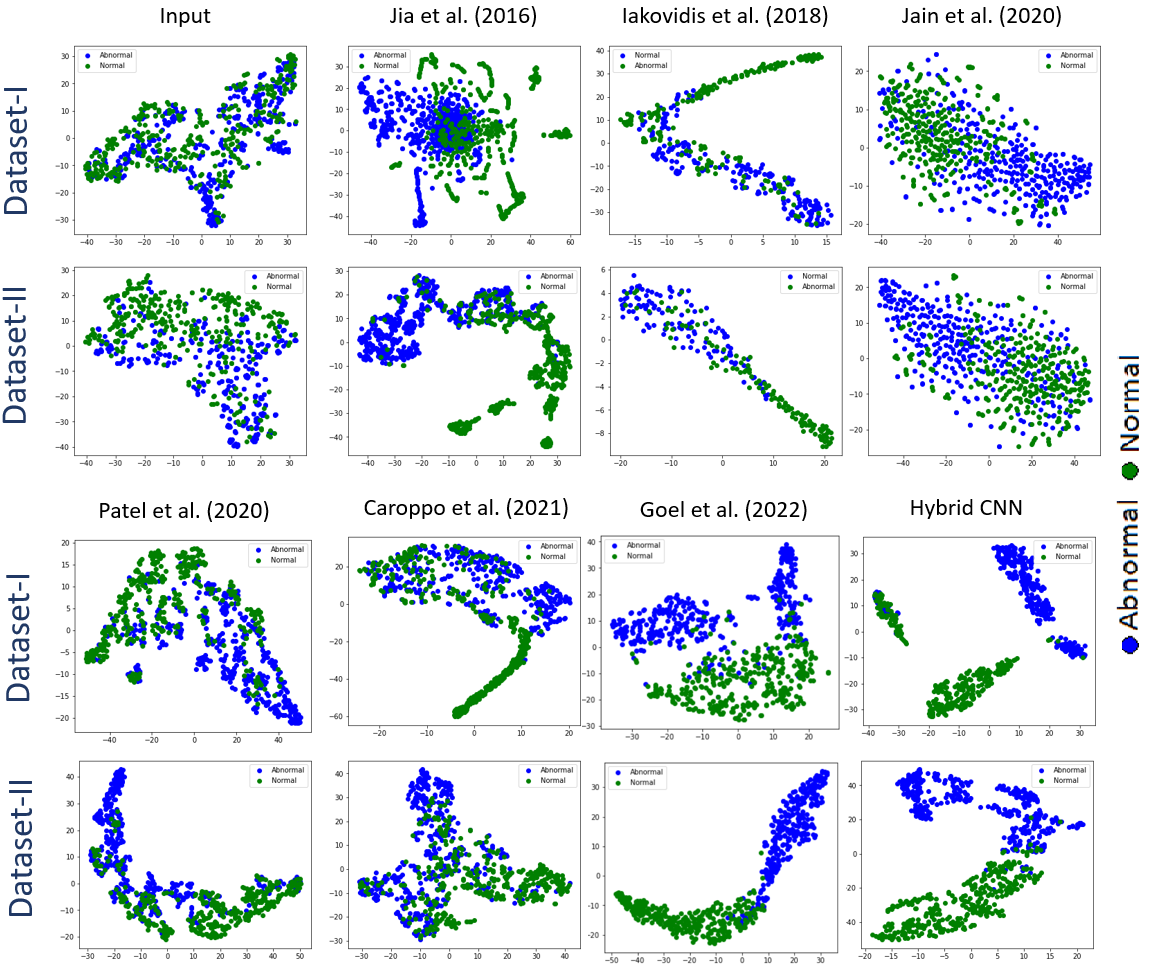}
	\centering
	\tiny
	\caption{T-SNE plots visualizing feature distribution of state-of-the-art methods and the hybrid CNN.}
	\label{tsne}
\end{figure}

To further illustrate the distinctive feature learning capability of the proposed CNN, the probability distributions of extracted features exploited for the classification of input images are visualized in Fig.\ref{tsne}. The feature distribution is plotted with the help of t-distributed stochastic neighbor embedding (t-SNE) \cite{cao2021epileptic}. T-SNE is a non-linear algorithm that reduces data dimensionality by considering the joint probabilities. It minimizes the Kullback-Leibler divergence of a low-dimensional embedding of high-dimensional data. The t-SNE method is utilized for mapping the probability feature vectors to the 2-dimensional space. The low dimensional feature embedding of the features obtained by the proposed hybrid CNN and those of six state-of-the-art methods are compared as shown in Fig.\ref{tsne}. It is evident from the visualization that the features extracted by the hybrid CNN are well differentiated as compared to other methods. There is quite a small overlap in the spatial distributions of the features obtained by the hybrid CNN as compared to those of the previous methods.

\begin{figure}
	\includegraphics[scale=0.6]{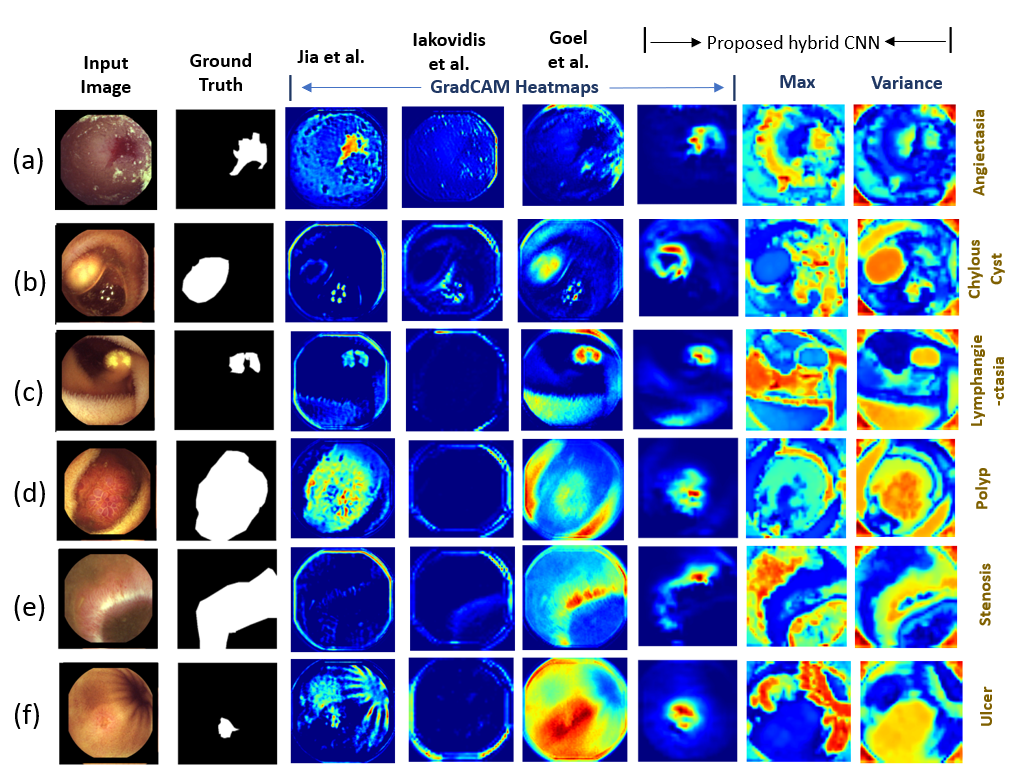}
	\centering
	\tiny
	\caption{Heatmaps acquired for the identification of salient regions using GradCAM with different CNN based state-of-the-art methods.}
	\label{gcam}
\end{figure}

\begin{table*}[!h]
	\centering
	\caption{Performance comparison of the proposed CNN with other state-of-the-art methods.}
	\label{SOTA}
	\scalebox{0.8}{\begin{tabular}{lccccccccccccc}  \hline
			& 
			\multicolumn{6}{c}{{Dataset-I}}&
			\multicolumn{6}{c}{{Dataset-II}}\\ 
			\textbf{Method}& \textbf{Type}  & \textbf{AUC} &\textbf{AC} & \textbf{PR} & \textbf{RE}  & \textbf{F1} & \textbf{CKS}&\textbf{AUC} &\textbf{AC} & \textbf{PR} & \textbf{RE}  & \textbf{F1} & \textbf{CKS}\\  \hline 
			Jia et al. \cite{jia2016deep} (2016) & DL &0.90&0.89 & 0.92 &0.88&0.90&0.879 & 0.91 & 0.90 &0.87& 0.92&0.89&0.924\\
			Iakovidis et al. \cite{iakovidis2018detecting} (2018)& DL & 0.92 &0.90& 0.91 &0.92 & 0.92 & 0.887& 0.93& 0.91 &0.89& 0.91&0.90& 0.928\\
			Patel et al. \cite{patel2020automated} (2020)& ML & 0.79&0.78 & 0.80 &0.78 & 0.79 & 0.481&0.85&0.82 &0.87& 0.79&0.73&0.567\\
			
			Jain et al. \cite{jain2020detection} (2020)& ML  & 0.91&0.88 & 0.89 &0.89 & 0.89 & 0.796&0.92&0.90 &0.90& 0.90&0.90&0.840\\
			
			Caroppo et al. \cite{caroppo2021deep} (2021)& DL+ML & \color{blue}{0.95}& 0.94 & 0.94 &0.94 & 0.94 & 0.901&0.96&0.95 &0.96& 0.93&0.95&0.932\\
			
			Goel et al. \cite{goel2022dilated} (2022)& DL+ML & 0.94& \color{blue}{0.95} & \color{blue}0.96 &0.93 & 0.94 & 0.916&0.93&0.93 &0.93& 0.93&0.93&0.856\\
			
			Hybrid CNN & DL & \color{blue}{0.95}&\color{blue}{0.95} & 0.95 & \color{blue}{0.96} &\color{blue}{0.95} & \color{blue}{0.921}&\color{blue}{0.97}&\color{blue}{0.97} &\color{blue}{0.97}&\color{blue}{0.97}&\color{blue}{0.97}&\color{blue}{0.948}\\
			
			Hybrid CNN + RF & DL+ML& \color{brown}\textbf{0.96}&\color{brown}\textbf{0.97} &\color{brown} \textbf{0.98} &\color{brown}\textbf{0.98} &\color{brown} \textbf{0.98} &\color{brown} \textbf{0.931}&\color{brown}\textbf{0.98}&\color{brown} \textbf{0.98} &\color{brown}\textbf{0.98}&\color{brown} \textbf{0.99}&\color{brown}\textbf{0.98}&\color{brown}\textbf{0.956}\\
			
			\hline
	\end{tabular}}
	\begin{itemize}
	\centering
	\item \scriptsize{Here \textcolor{brown}{brown} highlights the best results and \textcolor{blue}{blue} points out the second best results.}
	\end{itemize}
\end{table*} 
It may be noted that the methods developed over the DL platform \cite{jia2016deep, iakovidis2018detecting, caroppo2021deep, goel2022dilated} perform better than the conventional ML methods. A qualitative analysis of the DL models is also performed using the heatmaps of the features obtained from the last Conv layer of each of the CNN model. The heatmaps shown in Fig. \ref{gcam} are generated by the gradient-weighted class activation mapping (GradCAM) method \cite{jain2021deep}. As the Dataset-I contain ground truths in contrast to the Dataset-II, heatmaps are generated for six randomly selected samples from the Dataset-I with different types of anomalies namely angiectasia, chylous cyst, lymphangiectasia, polyp, stenosis, and ulcer. Heatmaps generated from the method by Jia et al. \cite{jia2016deep} indicate that the network has learnt to identify the vascular regions very well (row (a) and (d)) whereas the dilated CNN proposed by Goel et al. \cite{goel2022dilated} misses the vascular locations in the sample (a) and predicts the same with a small probability for the sample (d). It can be due to the limitations of dilated Conv in understanding local features and small size of anomalies \cite{hamaguchi2018effective}. In the sample (f), the CNN by Jia et al.\cite{jia2016deep} is misled by the dark normal region and identifies it as an anomaly. This could be since that their model is a shallow network that probably does not learn the finer textural details. In all these examples the proposed hybrid-CNN is performing better where the false positives are very less as compared to the other methods. It is worthwhile to mention here that the SFMs obtained from the hybrid CNN plotted in the last two columns in Fig. \ref{gcam} indicate the advantage of the meta-feature learning process of the proposed model. It can be seen that the max feature map highlights the normal regions whereas the variance emphasizes the abnormal regions in most of the cases. Surprisingly, they complement each other in the feature learning process leading to a powerful CNN model.               
 
\begin{figure}
	\includegraphics[scale=0.6]{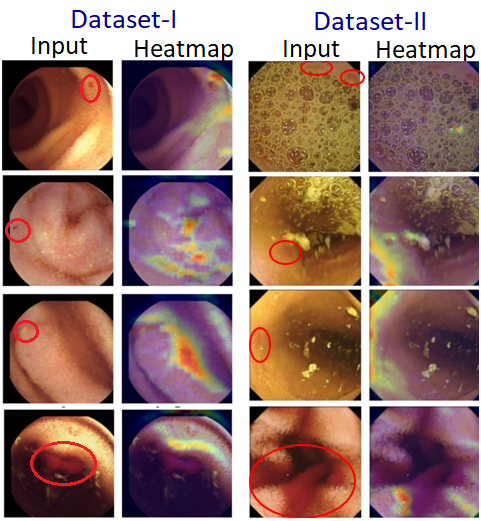}
	\centering
	\tiny
	\caption{Samples of WCE abnormal images falsely classified as normal by the hybrid CNN.}
	\label{chal}
\end{figure} 

Although the hybrid CNN shows better performance than other state-of-the-art methods, there are cases where it struggles to find the correct class of the image. Some challenging cases are displayed in Fig.\ref{chal}. It can be perceived from the results that if the size of the anomaly is very small or if there are artifacts like bubbles, specularity, or turbidity in the frame, the proposed CNN may find it difficult to identify the correct class. Real world applications of any deep learning model demand robustness against variations in the unseen samples. While the model should exhibit high precision, the recall rate or model's sensitivity depicts the robustness of the model \cite{yang2019ai}. The proposed model shows a remarkable performance with the highest recall of 98\% and 99\% on Dataset-I and II respectively as reported in Table. \ref{SOTA} respectively.

In the real world scenario, the proposed CNN may also be utilized for identifying abnormal frames in videos taken from a different apparatus. To illustrate the interoperability of the proposed method, symptomatic experiments were performed for cross-dataset validation of all the models considered for comparison. Performance of the models trained on Dataset-I and Dataset-II were tested on the CVC-CLINIC dataset \cite{bernal2015wm} that contains 612 image frames of 31 different polyps taken from various videos. The dataset contains only abnormal images. The results listed in Table \ref{sota_polyp} show that DL methods perform better due to their representation learning capability. The proposed CNN outperforms other methods with the highest RE of 87\% and 79\% on Dataset- I and Dataset- II respectively. This shows its suitability for real world applications as the hybrid CNN learns a rich pool of diverse features using the statistical features of the feature maps. 
\begin{table}[]
	\centering
	\caption{Comparative analysis for interoperability of pre-trained models from state-of-the-art methods and proposed hybrid-CNN on unseen CVC-CLINIC dataset.}
	\label{sota_polyp}
	\scalebox{0.8}{\begin{tabular}{lcccccc} \hline
			& 
			\multicolumn{3}{c}{{Dataset-I}}&
			\multicolumn{3}{c}{{Dataset-II}}\\ 
			\textbf{Method}& 
			\textbf{TP}  & \textbf{FN} & \textbf{RE}& \textbf{TP}  &\textbf{FN} & \textbf{RE}  \\	\hline 
			Jia et al. \cite{jia2016deep} (2016) & 487 &125 & 0.79 &422 & 190 & 0.68 \\
			Iakovidis et al. \cite{iakovidis2018detecting} (2018)& 449 & 163 & 0.73 &454 & 158 & 0.74 \\
			Patel et al. \cite{patel2020automated} (2020)& 240 & 372 & 0.39 &316 & 296 & 0.52\\
			Jain et al. \cite{jain2020detection} (2020)& 336  & 276 & 0.55 & 292 & 320 & 0.48\\
			Caroppo et al. \cite{caroppo2021deep} (2021)& 483 & 129 & 0.78 & 460 & 152 & 0.75\\
			Goel et al. \cite{goel2022dilated} (2022)& 276 & 336 & 0.45 & 439 & 173 & 0.71\\
			Hybrid CNN + RF & \textbf{535} & \textbf{77} & \textbf{0.87} & \textbf{482} & \textbf{130} & \textbf{0.79} \\
			\hline
	\end{tabular}}
\end{table} 

In addition to CVC-CLINIC dataset, cross-dataset validation is also performed on Dataset-I and Dataset-II. The models trained on Dataset-I are tested on Dataset-II and vice versa. The results are manifested in Table \ref{SOTA-CROSS}. It can be seen that the proposed method performs better with the highest F1-score 60\% on Dataset-II. In case of Dataset-I although it achieves the highest PR of 86\% , the overall performance of the hybrid CNN is slightly lower as the F1 score goes down to 62\% that is 2\% less than the dilated CNN \cite{goel2022dilated}. It can be deduced from the results that none of the methods outperform the others with respect to all the measures, but the proposed hybrid CNN is performing relatively better. The cross-validation results are not so impressive, probably due to the dissimilarity in images of both the datasets (please refer Fig. \ref{fanomaly}). The study presents that despite the availability of very efficient feature learning models, there is a need to further interpret the possible reasons for the under-performance of the DL methods. Large datasets with a wide variety of anomalies can help in improving a model's learning performance.

\begin{table}[]
	\centering
	\caption{Cross dataset validation results of the proposed CNN and other state-of-the-art methods on dataset-I and dataset-II }
	\label{SOTA-CROSS}
	\scalebox{0.8}{\begin{tabular}{lcccccccc} \hline
			& 
			\multicolumn{4}{c}{{Dataset-I}}&
			\multicolumn{4}{c}{{Dataset-II}}\\ 
			\textbf{Method}&  \textbf{AC} & \textbf{PR} & \textbf{RE}  & \textbf{F1} & \textbf{AC} & \textbf{PR} & \textbf{RE}  & \textbf{F1}\\  \hline 
			Jia et al. \cite{jia2016deep} (2016) & \textbf{0.51} &0.79&0.49 & 0.60 &0.48&\textbf{0.78}&0.49 &\textbf{0.60}\\
			Iakovidis et al. \cite{iakovidis2018detecting} (2018)& 0.45 & 0.81 &0.43& 0.56 &0.48 & 0.74 & 0.48& 0.58\\
			Patel et al. \cite{patel2020automated} (2020)& 0.49 & 0.62&0.49 & 0.55 &0.41 & 0.57 & 0.50&0.53\\
			
			Jain et al. \cite{jain2020detection} (2020)& 0.48& 0.79&0.51 & 0.62 &0.42 & 0.55 & 0.40&0.46\\
			
			Caroppo et al. \cite{caroppo2021deep} (2021)& 0.43 & 0.72& 0.47 & 0.56 &0.53 & 0.62 & 0.31&0.41\\
			
			Goel et al. \cite{goel2022dilated} (2022)& 0.45 & 0.79& \textbf{0.55} & \textbf{0.64} &\textbf{0.56} & 0.43 & 0.49&0.43\\
			
			Hybrid CNN + RF & 0.46& \textbf{0.86}&0.48 & 0.62 &\textbf{0.56} & 0.66& \textbf{0.55}&\textbf{0.60}\\
			
			\hline
	\end{tabular}}
\end{table} 
\section{Conclusion}
\color{black}
In this study, a hybrid CNN is proposed for abnormality detection in WCE images. It generates a rich pool of textural and statistical features that helps in the classification of normal and abnormal frames. CNC, DSC, and a novel concept of MFS layer is proposed for the extraction of features by the CNN trio. Two datasets namely, KID and Kvasir-Capsule have been used for model building and experimentation. Results show that the proposed CNN performs better than six state-of-the-art methods with $97\%$ and $98\%$ accuracy on KID and Kvasir-Capsule datasets, respectively. Experiments are also performed using the hybrid CNN as a feature extractor and it is observed that RF classifier combined with the proposed CNN as the feature extractor exhibits the best performance. Despite an impressive performance by the proposed model, there are certain limitations that the model faces. The issues with the artifacts like bubbles etc, that cause the false prediction are of concern. There is a need of large datasets with a wide variety of anomalies to improve the diagnostic performance of CNN models. Further, novel methods of training the network like the emerging concept of meta-learning can also be explored in this domain.          
\bibliographystyle{IEEEtran}  
\bibliography{ref}
\end{document}